\documentclass[10pt,twocolumn,letterpaper]{article}

\usepackage{iccv}
\usepackage{times}
\usepackage{epsfig}
\usepackage{graphicx}
\usepackage{amsmath}
\usepackage{amssymb}
\usepackage{enumitem}
\usepackage{siunitx}
\usepackage{booktabs}

\usepackage[breaklinks=true,bookmarks=false]{hyperref}

\iccvfinalcopy 

\def\iccvPaperID{****} 

\begin{document}

\title{Bag of Freebies for Training Object Detection Neural Networks}

\author{Zhi Zhang, Tong He, Hang Zhang, Zhongyue Zhang, Junyuan Xie, Mu Li\\
Amazon Web Services\\
{\tt\small \{zhiz, htong, hzaws, zhongyue, junyuanx, mli\}@amazon.com}
}

\maketitle

\begin{abstract}
Training heuristics greatly improve various image classification model accuracies~\cite{he2018bag}. Object detection models, however, have more complex neural network structures and optimization targets. The training strategies and pipelines dramatically vary among different models. In this works, we explore training tweaks that apply to various models including Faster R-CNN and YOLOv3. These tweaks do not change the model architectures, therefore, the inference costs remain the same. Our empirical results demonstrate that, however, these freebies can improve up to 5\% absolute precision compared to state-of-the-art baselines. 
\end{abstract}

\section{Introduction}

Object detection is no doubt one of the most fundamental applications in computer vision drawing attentions of researchers from various fields. Latest state-of-the-art detectors, including single (SSD \cite{liu2016ssd} and YOLO \cite{redmon2018yolov3}) or multiple stage RCNN-like \cite{girshick2015fast} neural networks and many variations or extended work  \cite{ren2015faster,Shen_2017}, are based on image classification backbone networks, e.g., VGG \cite{simonyan2014very}, ResNet \cite{he2016deep}, Inception \cite{szegedy2016rethinking} and MobileNet series \cite{howard2017mobilenets, sandler2018mobilenetv2}. 
Despite the rapid development and great success of the modern object detectors, different work usually employs different data prepossessing and training pipeline, which makes it hard for different object detection method to benefit from each other or relevant advancement in other area.  



\begin{figure}[t!]
  \centering
    \includegraphics[width=0.99\linewidth]{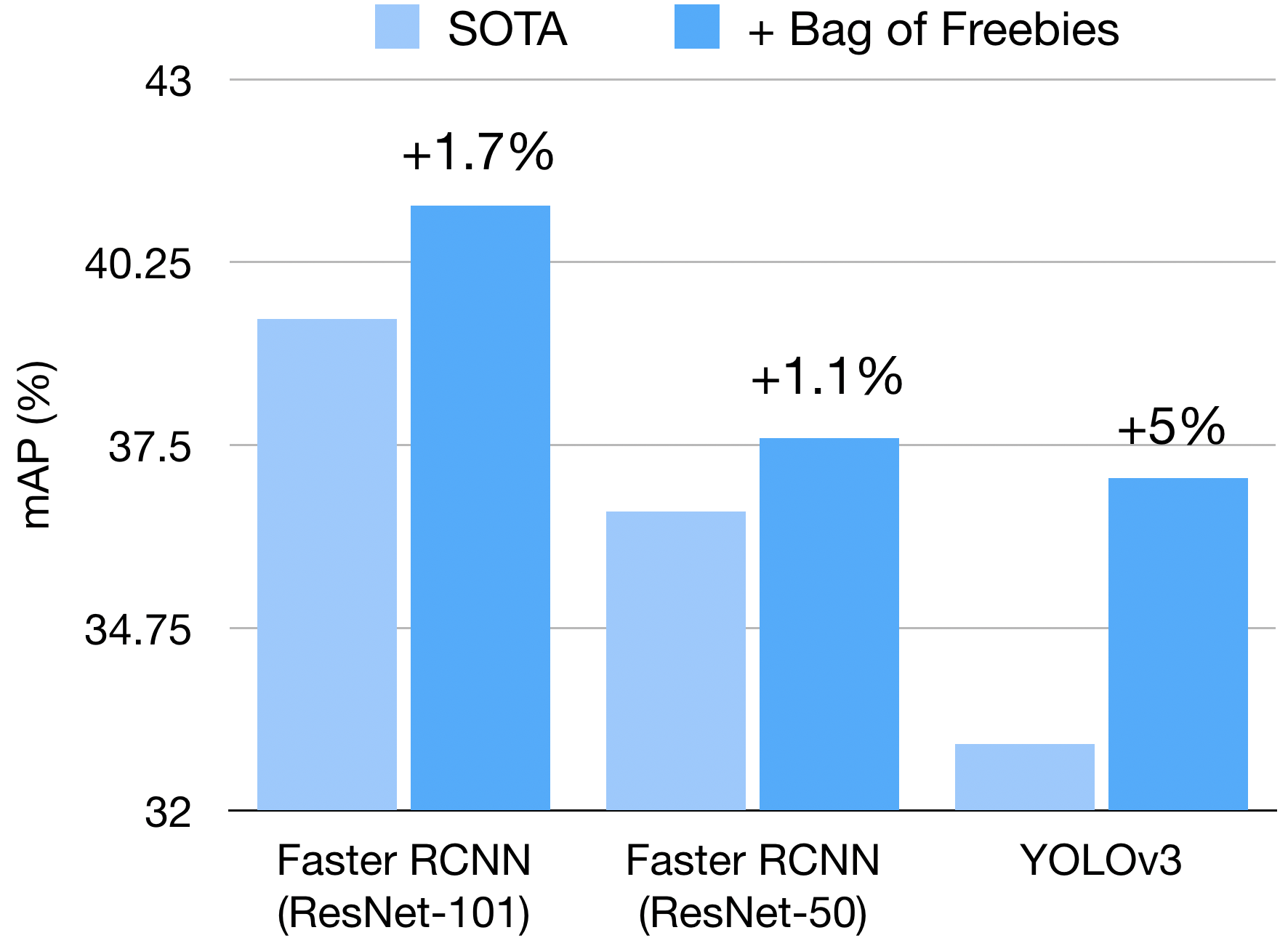}
  \caption{The Bag of Freebies improves object detector performances. There is no extra inference cost since models are not changed.}
  \label{fig:highlights}
\end{figure}

In this work, we focus on exploring effective and general approaches that can boost the performance of all popular object detection networks without introducing extra computational cost during inference. We first explore the mixup technique on object detection. Unlike \cite{zhang2017mixup}, we recognize the special property of multiple object detection task which favors spatial preserving transforms. Therefore we proposed a visually coherent image mixup methods designed for object detection tasks. Second, we explore detailed training pipelines including learning rate scheduling, label smoothing and synchronized BatchNorm~\cite{Zhang_2018_CVPR,peng2018megdet}. Third, we investigate the effectiveness of our training tweaks by incrementally stacking them to train single and multiple stage object detection networks.

Our major contributions can be summarized as follows:

\begin{enumerate}[itemsep=0mm]
    \item Systematically evaluating the various training heuristics applied to different object detection pipelines, providing valuable practice guidelines for future researches. To our best knowledge, this is the first work for surveying training heuristics for object detection.
    \item A visually coherent image mixup method designed for training object detection networks. Empirical results show that it is effective in improving model generalization capabilities.
    \item Extending the research depth on object detection data augmentation domain that strengthen the model generalization capability and help reduce over-fitting problems. 
    \item We achieve up to 5\% absolute precision improvement (15 to 20\% better than baseline) without modifying the network architectures. These model improvements bring no extra inference cost.
\end{enumerate}

The rest of this paper is organized as follows. First, we briefly introduce previous works in Section \ref{sec:related} on improving image classification and the potential to transfer to object detection models. Second, the proposed tweaks are detailed in Section \ref{sec:details}. Third, the experimental results will be benchmarked in Section \ref{sec:exp}. Finally, Section \ref{sec:conclusion} will conclude this work.

All related code are open-sourced and pre-trained weights for the models are available in GluonCV toolkit \cite{gluoncv}.

\section{Related Work}
\label{sec:related}
In this section, we briefly discuss related work regarding bag of tricks for image classification and heuristic object detection in common.

\subsection{Scattering tricks from Image Classification}
Image classification serves as the foundation of major 
computer vision tasks. Classification models are less computation intensive 
comparing with popular object detection and semantic segmentation models, therefore attractive enormous researchers to prototyping ideas. In this section, we briefly describe previous works that open the shed for this area. Learning rate warmup heuristic \cite{goyal2017accurate} is introduced to overcome the negative effect of extremely large mini-batch size. Interestingly, even though mini-batch size used in typical object detection training is nowhere close to the scale in image classification(\eg 10k or 30k \cite{goyal2017accurate}), a large amount of anchor size(up to 30k) is effectively contributing to batch size implicitly. A gradual warmup heuristic is crucial to YOLOv3 \cite{redmon2018yolov3} as in our experiments. There is a line of approaches trying to address the vulnerability of deep neural network. Label smoothing was introduced in \cite{szegedy2016rethinking}, which modifies the hard ground truth labeling in cross entropy loss. Zhang \etal \cite{zhang2017mixup} proposed \textit{mixup} to alleviate adversarial perturbation. Cosine annealing strategy for learning rate decay is proposed in \cite{loshchilov2016sgdr} in response to traditional step policy. He \etal achieved significant improvements on training accuracy by exploring bag of tricks \cite{he2018bag}. In this work, we dive deeper into the heuristic techniques introduced by image classification in the context of object detection. 

\subsection{Deep Object Detection Pipelines}
Most state-of-the-art deep neural network based object detection models are derived from multiple stages and single stage pipelines, starting from R-CNN \cite{girshick2014rich} and YOLO \cite{redmon2016you}, respectively. In single stage pipelines, predictions are generated by a single convolutional network and therefore preserve the spatial alignments (except that YOLO used Fully Connected layers at the end). However, in multiple stage pipelines, e.g. Fast R-CNN \cite{girshick2015fast} and Faster-RCNN \cite{ren2015faster}, final predictions are generated from features which are sampled and pooled in a specific region of interests (RoIs). RoIs are either propagated by neural networks or deterministic algorithms (e.g. Selective Search \cite{uijlings2013selective}). This major difference caused significant divergence in data processing and network optimization. For example, due to the lack of spatial variation in single stage pipelines, spatial data  augmentation is crucial to the performance as proven in Single-Shot MultiBox Object Detector (SSD) \cite{liu2016ssd}. Due to lack of exploration, many training details are exclusive to one series. In this work, we systematically explore the mutually beneficial tweaks and tricks that may help to boost the performance for both pipelines.

\section{Bag of Freebies}

\begin{figure*}[t!]
  \centering
    \includegraphics[width=0.85\linewidth]{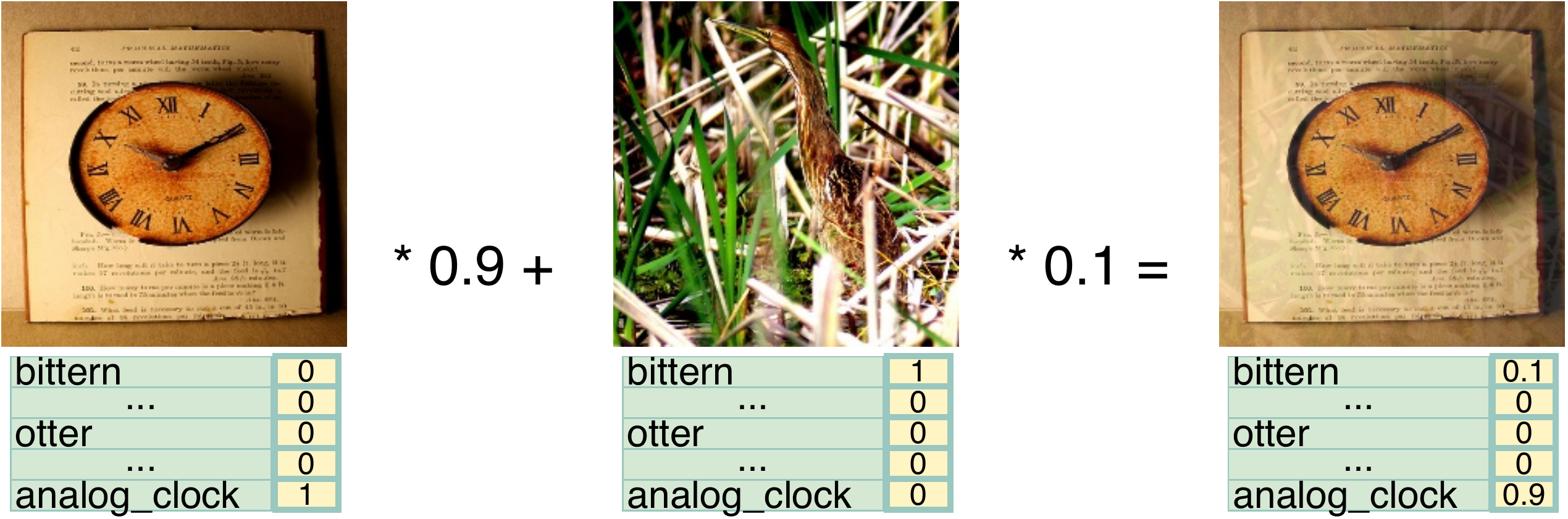}
  \caption{Mixup visualization of image classification with typical mixup ratio at $0.1:0.9$. Two images are mixed uniformly across all pixels, and image labels are weighted summation of original one-hot label vector. }
  \label{fig:classification_mixup}
\end{figure*}

\begin{figure*}[t!]
  \centering
    \includegraphics[width=0.8\linewidth]{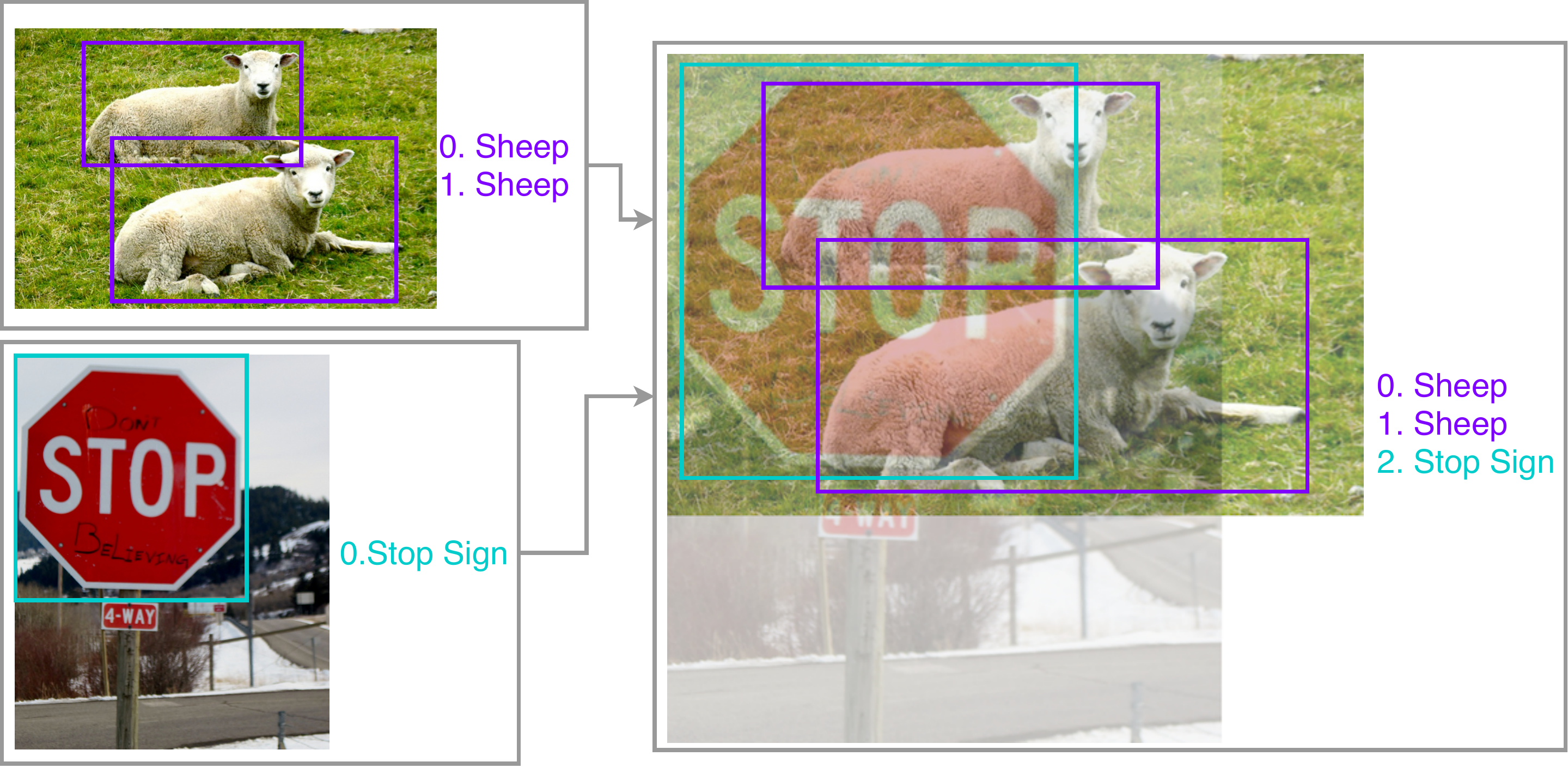}
  \caption{Geometry preserved alignment of mixed images for object detection. Image pixels are mixed up, object labels are merged as a new array.}
  \label{fig:det_mixup}
\end{figure*}

\label{sec:details}
In this section, we propose a visual coherent image mixup method for object detection. We will introduce data processing and training schedule designed to improve performance of object detection models.

\subsection{Visually Coherent Image Mixup for Object Detection}

\begin{figure}[t!]
  \centering
    \includegraphics[width=0.9\linewidth]{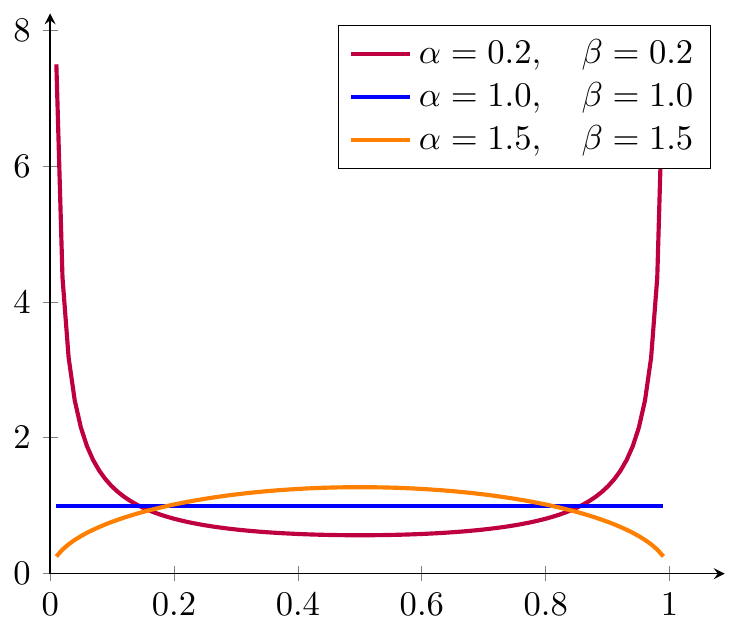}
  \caption{Comparison of different random weighted mixup sampling distributions. Red curve $\mathbf{B}(0.2,0.2)$ indicate the typical mixup ratio used in image classification. Blue curve is the special case $\mathbf{B}(1,1)$, equivalent to uniform distribution. Orange curve represents our choice $\mathbf{B}(1.5,1.5)$ for object detection after preliminary experiments. }
  \label{fig:beta_dist}
\end{figure}

\textit{Mixup} has been proved successful in alleviating adversarial perturbations in classification networks~\cite{zhang2017mixup}. The key idea of mixup in image classification task is to regularize the neural network to favor simple linear behavior by mixing up pixels as interpolations between pairs of training images. At the same time, one-hot image labels are mixed up using the same ratio. An example of mixup in image classification is illustrated in Fig.~\ref{fig:classification_mixup}. 

The distribution of blending ratio in mixup algorithm proposed by Zhang \etal \cite{zhang2017mixup} is drawn from a beta distribution $\mathbf{B}(0.2,0.2)$. 
The majority of mixups are barely noises with such beta distributions. Rosenfeld \etal~\cite{rosenfeld2018elephant} conduct a series of interesting experiments named as ``Elephant in the room", where an elephant patch is randomly placed on a natural image, then this adversarial image is used to challenge existing object detection models. The results indicate that existing object detection models are prune to such attack and show weakness to detect such transplanted objects.

Inspired by the heuristic experiments by Rosenfeld \etal \cite{rosenfeld2018elephant}, we focus on the natural co-occurrence object presentations which play significant roles in object detection. By applying more complex spatial transforms, we introduce occlusions, spatial signal perturbations that are common in natural image presentations. 

In our empirical experiments, continue increasing blending ratio used in the mixup process, the objects in resulting frames are more vibrant and coherent to the natural presentations, similar to the transition frames commonly observed when we are watching low FPS movies or surveillance videos. The visual comparisons of image classification and such high ratio mixup are illustrated in Fig.~\ref{fig:classification_mixup} and Fig.~\ref{fig:det_mixup}, respectively. In particular, we use geometry preserved alignment for image mixup to avoid distort images at the initial steps. We also choose a beta distribution with $\alpha$ and $\beta$ are both at least 1, which is more visually coherent, instead of following the same practice in image classification, as depicted in Figure \ref{fig:beta_dist}. 

\begin{table}[t!]
\begin{center}
\begin{tabular}{l|c}
Model                           & mAP @ 0.5 \\
\specialrule{1pt}{1pt}{1pt}
baseline                   & 81.5      \\
0.5:0.5 evenly           & 83.05     \\
$\mathbf{B}(1.0, 1.0)$, weighted loss & 83.48     \\
$\mathbf{B}(1.5, 1.5)$, weighted loss & 83.54    
\end{tabular}
\end{center}
\caption{Effect of various mixup approaches, validated with YOLOv3 \cite{redmon2018yolov3} on Pascal VOC 2007 test set. \textbf{Weighted loss} indicates the overall loss is the summation of multiple objects with ratio 0 to 1 according to image blending ratio they belong to in the original training images. }
\label{tab:yolo-mixup}
\end{table}

To verify mixup designed for object detection, we experimentally tested empirical mixup ratio distributions using the YOLOv3 network on Pascal VOC dataset. Table.~\ref{tab:yolo-mixup} shows the actual improvements by adopting detection mixups with ratios sampled by different beta distributions. Beta distribution with $\alpha$ and $\beta$ both equal to 1.5 is marginally better than 1.0 (equivalent to uniform distribution) and better than fixed even mixup. We recognize that for object detection where mutual object occlusion is common, networks are encouraged to observe unusual crowded patches, either presented naturally or created by adversarial techniques. 



\begin{figure*}[t!]
  \centering
    \includegraphics[width=0.99\linewidth]{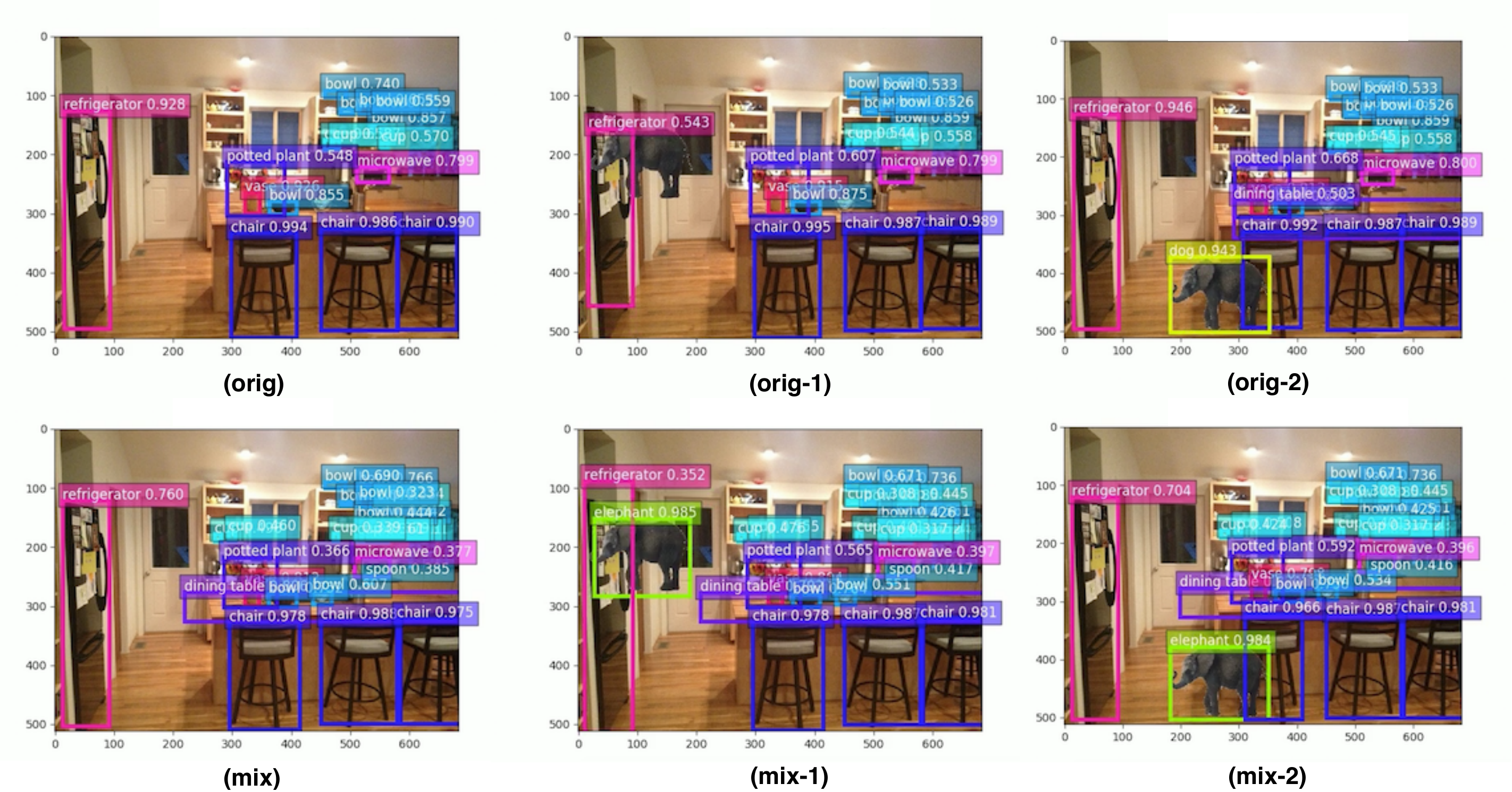}
  \caption{Elephant in the room example. Model trained with geometry preserved mixup (bottom) is more robust against alien objects compared to baseline (top).}
  \label{fig:mixup_elephant}
\end{figure*}

To validate the effectiveness of visually coherent mixup, we followed the same experiments of "Elephant in the room" \cite{rosenfeld2018elephant} by sliding an elephant image patch through an indoor room image. We trained two YOLOv3 models on COCO 2017 dataset with identical settings except for that model mix is using our mixup approach. We depict some surprising discoveries in Fig.~\ref{fig:mixup_elephant}. As we can observe in Fig.~\ref{fig:mixup_elephant}, vanilla model trained without our mix approach is struggles to detect "elephant in the room" due to heavy occlusion and lack of context because it's rare to capture an elephant in a kitchen. Actually, there is no such training image after examine the common training datasets. In comparison, models trained with our mix approach is more robust thanks to randomly generated visually deceptive training images. In addition, we also notice that mix model is more \textit{humble}, less confident and generates lower scores for objects on average. However, this behavior does not affect evaluation results as shown in experimental results. We evaluated the model performance against fake video with elephant sliding through, and the results are listed in Table.~\ref{tab:elephant-result}. It is obvious that model trained with visually coherent mixup is more robust (94.12 vs. 42.95) to detect elephant in indoor scene even though it is very rare in natural images. And mixup model can preserve crowded furniture objects under heavy occlusion of alien elephant image patch. We recognize that mixup model receives more challenges during training therefore is significantly better than vanilla model in handling unprecedented scenes and very crowded object groups.

\begin{table}[t!]
\begin{center}
\begin{tabular}{l|c|c}
Model                           & recall of elephant & disappeared furniture \\
\specialrule{1pt}{1pt}{1pt}
baseline                   & 42.95  &  8.24 \%  \\
+mixup           & \textbf{94.12} & \textbf{1.27\%}    
\end{tabular}
\end{center}
\caption{Statistics of detection results affected by elephant in the room. "Recall of elephant" is the recall of sliding elephant in all generated frames, indicating how robust the model handles objects in unseen context. Disappeared furniture percentage is calculated by dividing sum of disappeared furniture count by overall furniture objects in all adversarial frames.}
\label{tab:elephant-result}
\end{table}

\subsection{Classification Head Label Smoothing}
For each object, detection networks often compute a probability distribution over all classes with softmax function:

\begin{equation}
p_i = \frac{e^{z_i}}{\sum_j e^{z_j}},
\end{equation}

where $z_i$s are the unnormalized logits directly from the last linear layer for classification prediction. For object detection during training, we only modify the classification loss by comparing the output distribution $p$ against the ground truth distribution $q$ with cross-entropy

\begin{equation}
L = -\sum_i q_i \log p_i.
\end{equation}

$q$ is often a one-hot distribution, where the correct class has probability one while all other classes have zero. Softmax function, however, can only approach this distribution when $z_i \gg z_j, \forall j \neq i$ but never reach it. This encourages the model to be too confident in its predictions and is prone to over-fitting.

Label smoothing was proposed by Szegedy \etal~\cite{szegedy2016rethinking} as a form of regularization. We smooth the ground truth distribution with
\begin{equation}
q_i =
\begin{cases}
  1-\varepsilon & \quad\textrm{if } i = y, \\
   \varepsilon / (K-1) & \quad\textrm{otherwise,}\\
\end{cases}
\label{eq:label-smoothing}
\end{equation}
where $K$ is the total number of classes and $\varepsilon$ is a small constant. This technique reduces the model's confidence, measured by the difference between the largest and smallest logits.

In the case of sigmoid outputs of range \numrange{0}{1.0} as in YOLOv3~\cite{redmon2018yolov3}, label smoothing is even simpler by correcting the upper and lower limit of the range of targets as in Eq. \ref{eq:label-smoothing}.

\subsection{Data Preprocessing}
In image classification domain, usually neural networks are extremely tolerant to image geometrical transformation. It is actually encouraged to randomly perturb the spatial characteristics, \eg randomly flip, rotate and crop images in order to improve generalization accuracy and avoid overfitting. However, for object detection image preprocessing,  we need to carry additional cautious since detection networks are more sensitive to such transformations. 

We experimentally review the following data augmentation methods:

\begin{itemize}
   \item Random geometry transformation. Including random cropping (with constraints), random expansion, random horizontal flip and random resize (with random interpolation).
   \item Random color jittering including brightness, hue, saturation, and contrast.
\end{itemize}

In terms of types of detection networks, there are two pipelines for generating final predictions. First is single stage detector network, where final outputs are generated from every single cell in the feature map, for example SSD\cite{liu2016ssd} and YOLO\cite{redmon2018yolov3} networks which generate detection results proportional to spatial shape of an input image. The second is multi-stage proposal and sampling based approaches, following Fast-RCNN\cite{ren2015faster}, where a certain number of candidates are sampled from a large pool of generated ROIs, then the detection results are produced by repeatedly cropping the corresponding regions on feature maps, and the number of predictions is proportional to number of samples.

Since sampling-based approaches conduct enormous cropping operations on feature maps, it substitutes the operation of randomly cropping input images, therefore these networks do not require extensive geometric augmentations applied during the training stage. This is the major difference between one-stage and so called multi-stage object detection data pipelines. In our Faster-RCNN training, we do not use random cropping techniques during data augmentation.

\subsection{Training Schedule Revamping}
During training, the learning rate usually starts with a relatively big number and gradually becomes smaller throughout the training process. For example, the step schedule is the most widely used learning rate schedule. With step schedule, the learning rate is multiplied by a constant number below 1 after reaching pre-defined epochs or iterations. For instance, the default step learning rate schedule for Faster-RCNN \cite{ren2015faster} is to reduce learning rate by ratio $0.1$ at $60k$ iterations. Similarly, YOLOv3 \cite{redmon2018yolov3} uses same ratio $0.1$ to reduce learning rate at $40k$ and $45k$ iterations. Step schedule has sharp learning rate transition which may cause the optimizer to re-stabilize the learning momentum in the next few iterations. In contrast, a smoother cosine learning rate adjustment was proposed by Loshchilov \etal~\cite{loshchilov2016sgdr}. 
Cosine schedule scales the learning rate according to the value of cosine function on 0 to pi. It starts with slowly reducing large learning rate, then reduces the learning rate quickly halfway, and finally ends up with tiny slope reducing small learning rate until it reaches $0$. In our implementation, we follow He \etal \cite{he2018bag} but the numbers of iterations are adjusted according to object detection networks and datasets.

Warmup learning rate is another common strategy to avoid gradient explosion during the initial training iterations. Warmup learning rate schedule is critical to several object detection algorithms, e.g., YOLOv3, which has a dominant gradient from negative examples in the very beginning iterations where sigmoid classification score is initialized around 0.5 and biased towards 0 for the majority predictions.

\begin{figure}[t!]
  \centering
    \includegraphics[width=0.99\linewidth]{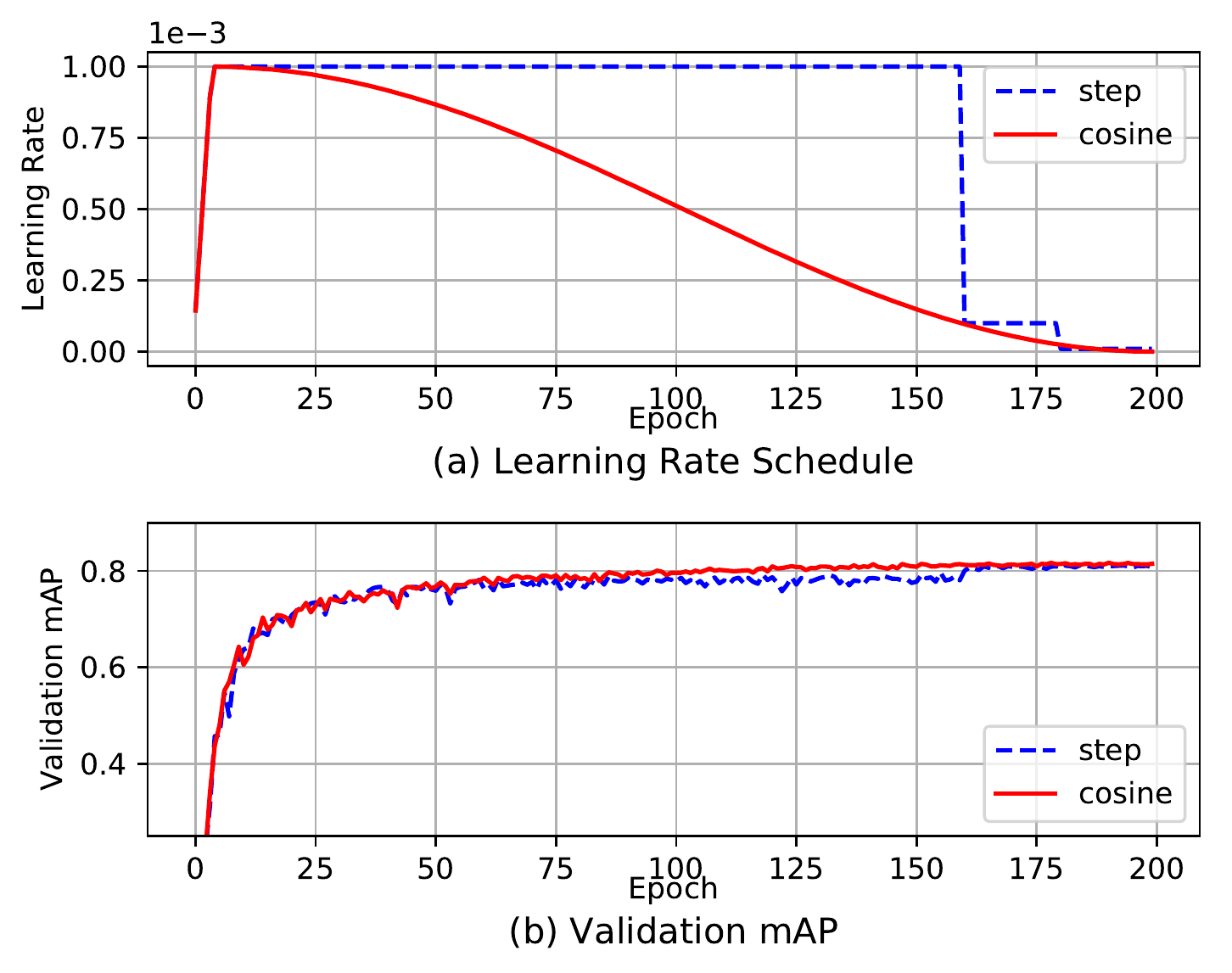}
  \caption{Visualization of learning rate scheduling with warmup enabled for YOLOv3 training on Pascal VOC. (a): cosine and step schedules for batch size 64. (b): Validation mAP comparison curves using step and cosine learning schedule.}
  \label{fig:lr_schedule}
\end{figure}

Training with cosine schedule and proper warmup lead to better validation accuracy, as depicted in Fig.~\ref{fig:lr_schedule}, validation mAP achieved by applying cosine learning rate decay outperforms step learning rate schedule at all times in training. Due to the higher frequency of learning rate adjustment, it also suffers less from plateau phenomenon of step decay that validation performance will be stuck for a while until learning rate is reduced. 

\subsection{Synchronized Batch Normalization}

In recent years, the need of massive computation resources forces training environments to equip multiple devices (usually GPUs) to accelerate training. Despite handling different hyper-parameters in response to larger batch sizes during training, Batch Normalization \cite{ioffe2015batch} is drawing the attention of multi-device users due to the implementation details. Although the typical implementation of Batch Normalization working on multiple devices (GPUs) is fast (with no communication overhead), it inevitably reduces the size of batch size and causing slightly different statistics during computation, which potentially degraded performance. This is not a significant issue in some standard vision tasks such as ImageNet classification (as the batch size per device is usually large enough to obtain good statistics). However, it hurts the performance in some tasks with a small batch size (\eg, 1 per GPU). Recently, Peng \etal~\cite{peng2018megdet} has proved the importance of synchronized batch normalization in object detection. 
In this work, we review the importance of Synchronized Batch Normalization with YOLOv3 \cite{redmon2018yolov3} to evaluate the impacts of relatively smaller batch-size on each GPU as training image shape is significantly larger than image classification tasks.



\subsection{Random shapes training for single-stage object detection networks}

Natural training images come in various shapes. To fit memory limitation and allow simpler batching, many single-stage object detection networks are trained with fixed shapes \cite{liu2016ssd, redmon2016you}. To reduce risk of overfitting and to improve generalization of network predictions, we follow the approach of random shapes training as in Redmon \etal \cite{redmon2018yolov3}. More specifically, a mini-batch of $N$ training images is resized to $N\times 3 \times H \times W$, where H and W are multipliers of network stride. For example, we use $H = W \in \{320, 352, 384, 416, 448, 480, 512, 544, 576, 608\}$ for YOLOv3 training given the stride of feature map is $32$.

\section{Experiments}
\label{sec:exp}

In order to compare proposed tweaks for object detection, we pick up one popular object detection framework from single and multiple stage pipelines, respectively. YOLOv3 \cite{redmon2018yolov3} is famous for its efficiency and good accuracy. Faster-RCNN \cite{ren2015faster} is one of the most adopted detection framework and foundation of many others variants. Therefore in this paper, we use YOLOv3 and Faster-RCNN as representatives to conduct experiments. Note that in order to remove side effects of test time tricks, we always report single scale, single model results with standard Non-maximum Suppression implementation. We do not use external training image or labels in our experiments.

\subsection{Incremental trick evaluation on Pascal VOC}
Pascal VOC is the most common dataset for benchmarking object detection models \cite{girshick2015fast,liu2016ssd,redmon2016you}, we use Pascal VOC 2007 \textit{trainval} and 2012 \textit{trainval} for training and 2007 test set for validation. The results are reported in mean average precision defined in Pascal VOC development kit \cite{everingham2010pascal}. For YOLOv3 models, we consistently validate mean average precision (mAP) at $416 \times 416$ resolution. If random shape training is enabled, YOLOv3 models will be fed with random resolutions from $320 \times 320$ to $608 \times 608$ with $32 \times 32$ increments, otherwise they are always trained with fixed $416 \times 416$ input data. Faster RCNN models take arbitrary input resolutions. In order to regulate training memory consumption, the shorter sides of input images are resized to $600$ pixels while ensuring the longer side in smaller than $1000$ pixels. Training and validation of Faster-RCNN models follow the same preprocessing steps, except that training images have chances of 0.5 to flip horizontally as additional data augmentation. The incremental evaluations of YOLOv3 and Faster-RCNN with our bags of freebies (BoF) are detailed in Table.~\ref{tab:yolo3-voc} and Table.~\ref{tab:frcnn-voc}, respectively. 

\begin{table}[t!]
\begin{center}
\begin{tabular}{l|c|c|c}

Incremental Tricks        & mAP &  $\Delta$ & Cumu $\Delta$ \\ \specialrule{1pt}{1pt}{1pt}
- data augmentation        &  64.26  & -15.99  & -15.99    \\ 
baseline                  &  80.25  & 0 & 0  \\
+ synchronize BN          &  80.81   & +0.56  & +0.56     \\
+ random training shapes  &  81.23  & +0.42 & +0.98       \\
+ cosine lr schedule        &  81.69  & +0.46  & +1.44  \\
+ class label smoothing     &  82.14  & +0.45  & +1.89  \\
+ mixup & \textbf{83.68}   & \textbf{+1.54} & \textbf{+3.43}
\end{tabular}
\end{center}
\caption{Incremental trick validation results of YOLOv3, evaluated at $416 \times 416$ on Pascal VOC 2007 test set. }
\label{tab:yolo3-voc}
\end{table}

\begin{table}[t!]
\begin{center}
\begin{tabular}{l|c|c|c}

Incremental Tricks         & mAP  & $\Delta$ & Cumu $\Delta$ \\ \specialrule{1pt}{1pt}{1pt}
- data augmentation        &    77.61  & -0.16 & -0.16    \\ 
baseline                  &   77.77  & 0  & 0 \\
+ cosine lr schedule        &   79.59  & +1.82  & +1.82 \\
+ class label smoothing           &  80.23  & +0.64 & +2.46  \\
+ mixup &   \textbf{81.32}  & \textbf{+0.89} & \textbf{+3.55}
\end{tabular}
\end{center}
\caption{Incremental trick validation results of Faster-RCNN, evaluated at $600 \times 1000$ on Pascal VOC 2007 test set. }
\label{tab:frcnn-voc}
\end{table}

\begin{table*}[t!]
\begin{center}
\begin{tabular}{l|c|c|c}

Model                       & Orig $mAP_{bbox}^{0.5:0.95}$ & \textbf{Our BoF $mAP_{bbox}^{0.5:0.95}$} & Absolute delta \\ 
\specialrule{1pt}{1pt}{1pt}
Faster-RCNN R50 \cite{Detectron2018}        &  36.5 & \textbf{37.6}  &  \textbf{+1.1}  \\ 
Faster-RCNN R101 \cite{Detectron2018}        &  39.4 & \textbf{41.1}  &  \textbf{+1.7}  \\ 
\hline
YOLOv3 @320 \cite{redmon2018yolov3}  &   28.2  & \textbf{33.6}     &  \textbf{+5.4} \\
YOLOv3 @416 \cite{redmon2018yolov3}  &   31.0  & \textbf{36.0}     &  \textbf{+5.0} \\
YOLOv3 @608 \cite{redmon2018yolov3}  &   33.0  & \textbf{37.0}     &  \textbf{+4.0} \\
\end{tabular}
\end{center}
\caption{Overview of improvements achieved by applying bag of freebies(BoF), evaluated on MS COCO \cite{lin2014microsoft} 2017 val set. Note that YOLOv3 models can be evaluated at different input resolutions with same weights, our BoF improves evaluation results more significantly at lower resolution levels.}
\label{tab:coco}
\end{table*}

\begin{figure*}[t!]
  \centering
    \includegraphics[width=0.99\linewidth]{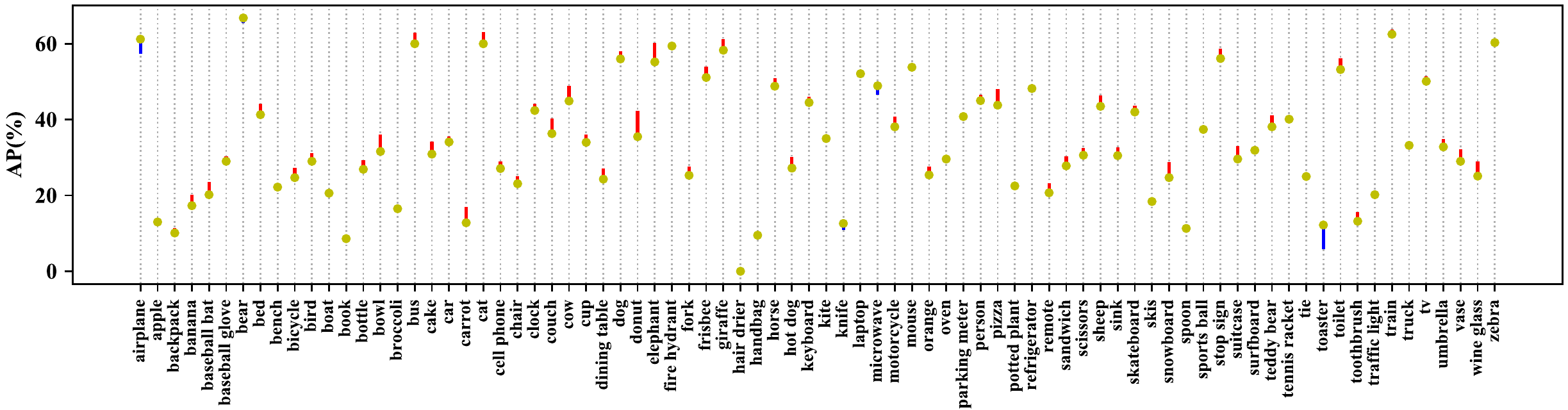}
  \caption{{COCO 80 category AP analysis with YOLOv3 \cite{redmon2018yolov3}.} Red lines indicate performance gain using BoF, while blue lines indicate performance drop. }
  \label{fig:yolo_coco_per_class}
\end{figure*}

\begin{figure*}[t!]
  \centering
    \includegraphics[width=0.99\linewidth]{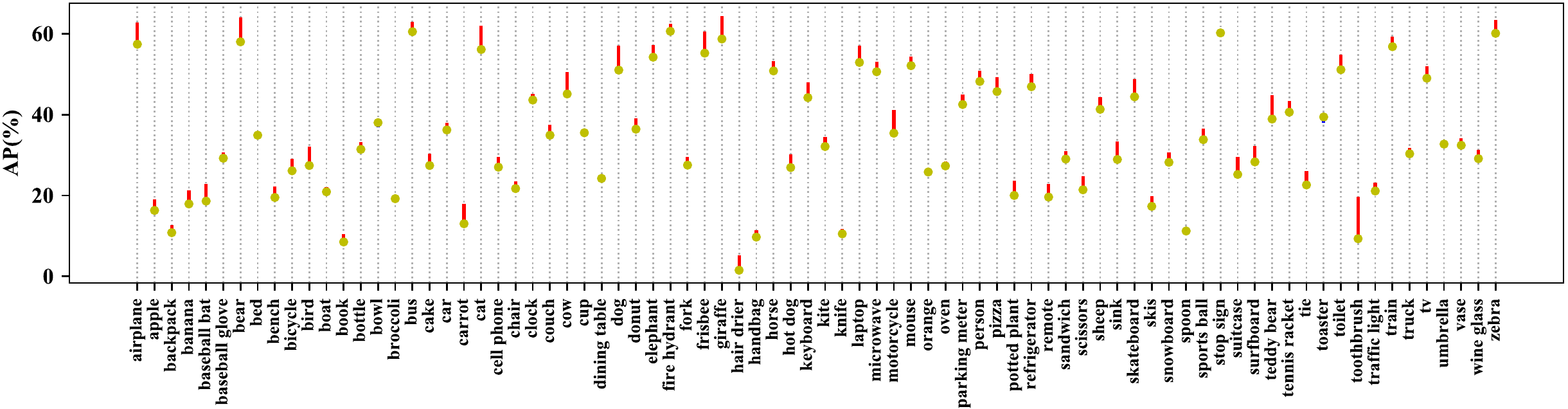}
  \caption{{COCO 80 category AP analysis with Faster-RCNN resnet50 \cite{ren2015faster}.} Red lines indicate performance gain using BoF, while blue lines indicate performance drop. }
  \label{fig:frcnn_coco_per_class}
\end{figure*}

For YOLOv3, we first notice that data augmentation contributed nearly 16\% to the baseline mAP, suggesting that single-stage object detection networks rely heavily on assistance of data augmentation to create unseen patches. In terms of the training tricks we mentioned in the previous section, stacking Synchronized BatchNorm, Random Training, cosine learning rate schedule, Sigmoid label smoothing and detection mixup continuously improves validation performance, up to 3.43\%, achieving 83.68\% single model single scale mAP.

For Faster-RCNN, one obvious difference compared with YOLOv3 results is that disabling data augmentation only introduced a minimal 0.16\% mAP loss. This phenomena is indicating that sampling based proposals can effectively replace random cropping which is heavily used in single stage object detection training pipelines. Second, incremental mAPs show strong confidence that the proposed tricks can effectively improve model performance, with a significant 3.55\% gain.

It is challenging to achieve mAP higher than 80\% with out external training data on Pascal VOC \cite{ren2015faster,liu2016ssd,Shen_2017}. However, we managed to achieve up to 3.5\% mAP gain on both YOLOv3 and Faster-RCNN models, reaching as high as 83.68\% single model single scale evaluation results.

\subsection{Bag of Freebies on MS COCO.}
To further evaluate effectiveness of bag of freebies on larger dataset, we benchmark on MS COCO \cite{lin2014microsoft} in order to validate the generalization of our bags of tricks in this work. COCO 2017 is 10 times larger than Pascal VOC and contains much more tiny objects compared with PASCAL VOC.  We use similar training and validation settings as Pascal VOC, except that Faster-RCNN models are resized to $800 \times 1300$ pixels in response to smaller objects. The results are shown in Table.~\ref{tab:coco}. 


\begin{table}[t!]
\begin{center}
\begin{tabular}{l|c|c}

                       &  -Mixup YOLO3 & +Mixup YOLO3  \\ \specialrule{1pt}{1pt}{1pt}
-Mixup darknet53        &  35.0    &   35.3   \\ 
+Mixup darknet53                 &   36.4  & 37.0   \\
\end{tabular}
\end{center}
\caption{Combined analysis of impacts of mixup methodology for pre-trained image classification and detection network.}
\label{tab:mixup-yolo3}
\end{table}

\begin{table}[t!]
\begin{center}
\begin{tabular}{l|c|c}

                       &  -Mixup FRCNN & +Mixup FRCNN \\ \specialrule{1pt}{1pt}{1pt}
-Mixup R101        &  39.9    &  40.1   \\ 
+Mixup R101                 &   40.1  & 41.1  \\
\end{tabular}
\end{center}
\caption{Combined analysis of impacts of mixup methodology for pre-trained image classification and detection network.}
\label{tab:mixup-frcnn101}
\end{table}

In summary, our proposed bags of freebies boost Faster-RCNN models by 1.1\% and 1.7\% absolute mean AP over existing state-of-the-art implementations \cite{Detectron2018} with ResNet 50 and 101 base models, respectively. Following evaluation resolution reports in \cite{redmon2018yolov3}, we list YOLOv3 evalution results using $320,416,608$ resolutions to compare performance at different scales. While at $608 \times 608$ our model outperforms baseline \cite{redmon2018yolov3} by 4.0\% absolute mAP, at lower resolutions, this gap is more significantly 5.4\% absolute mAP, almost 20\% better than baseline. 
Note that all these results are obtained by generating better weights in a fully compatible inference model, \ie, all these achievements are free lunch during inference. We also notice that by adopting bag of freebies during training, we successfully uplift YOLOv3 performance to the same level as state-of-the-art Faster-RCNN \cite{Detectron2018} (37.0 vs 36.5) while preserves faster inference speed as part of single stage model benefits. 

Mean AP is the average over 80 categories, which may not reflect the per category performance. We plot per category AP changes of YOLOv3 and Faster-RCNN models before and after our BoF in Fig.~\ref{fig:yolo_coco_per_class} and Fig.~\ref{fig:frcnn_coco_per_class} respectively. Except rare cases, we can see the majority of categories benefit from bag of freebies training tricks.

\subsection{Impact of mixup on different phases of training detection network}
Mixup can be applied in two phases of object detection networks: 1) pre-training classification network backbone with traditional mixup \cite{he2018bag, zhang2017mixup}; 2) training detection networks using proposed visually coherent image mixup for object detection. Since we do not freeze weights pre-trained on ImageNet, both training phase can affect final detection models. We compare the results using Darknet 53-layer based YOLO3 \cite{redmon2018yolov3} implementation and ResNet101 \cite{he2016deep} based Faster-RCNN \cite{ren2015faster}. Final validation results are listed in Table. \ref{tab:mixup-yolo3} and Table. \ref{tab:mixup-frcnn101}, respectively. While the results prove the consistent improvements by adopting mixup to either training phases, interestingly it is also notable that applying mixup in both phases can produce more significant gains. For example, employing either pre-training mixup or detection mixup has nearly 0.2\% absolute mAP improvement over baseline. By combining both mixup techniques, we achieve 1.2\% performance boost. We expect by applying mixup in both training phases, shallow layers of networks are receiving statistically similar inputs, resulting in less perturbations for low level filters. 

\section{Conclusion}
\label{sec:conclusion}

In this paper, we propose a bag of training enhancements significantly improved model performances while introducing zero overhead to the inference environment. Our empirical experiments of YOLOv3 \cite{redmon2018yolov3} and Faster-RCNN \cite{ren2015faster} on Pascal VOC and COCO datasets show that the bag of tricks are consistently improving object detection models. By stacking all these tweaks, we observe no signs of degradation of any level and suggest a wider adoption to future object detection training pipelines. These freebies are all training time modifications, and therefore only affect model weights without increasing inference time or change of network structures. All existing and future work will be included as part of open source GluonCV repository \cite{gluoncv}.

{\small
\bibliographystyle{ieee}
\bibliography{egbib}
}

\end{document}